\documentclass[11pt]{article}

\usepackage[preprint]{acl}

\usepackage{times}
\usepackage{latexsym}
\usepackage{amssymb} 
\usepackage{amsmath}
\usepackage[table]{xcolor}
\usepackage{tcolorbox}
\usepackage{multirow}
\usepackage{array}
\usepackage{booktabs}
\usepackage{makecell}

\usepackage[T1]{fontenc}

\usepackage[utf8]{inputenc}

\usepackage{microtype}

\usepackage{inconsolata}

\usepackage{graphicx}

\title{Prefix Probing: Lightweight Harmful Content Detection for Large Language Models}

\author{
    Jirui Yang$^{1}$,
    Hengqi Guo$^{1}$,
    Zhihui Lu$^{1}$,
    Yi Zhao$^{1}$,
    Yuansen Zhang$^{2}$\\
    \textbf{Shijing Hu$^{1}$,
    Qiang Duan$^{3}$,
    Yinggui Wang$^{2}$,
    Tao Wei$^{2}$}\\[0.3em]
    $^{1}$Fudan University \quad
    $^{2}$Ant Group \quad
    $^{3}$Pennsylvania State University\\
    \texttt{yangjr23@m.fudan.edu.cn,hqguo22@m.fudan.edu.cn,lzh@fudan.edu.cn}\\
    \texttt{zhaoyi25@m.fudan.edu.cn,yuansen.zys@antgroup.com,sjhu24@m.fudan.edu.cn}\\
    \texttt{qduan@psu.edu,wyinggui@gmail.com,lenx.wei@antgroup.com}
}

\begin{document}
\maketitle
\begin{abstract}

Large language models often face a three-way trade-off among detection accuracy, inference latency, and deployment cost when used in real-world safety-sensitive applications. This paper introduces Prefix Probing, a black-box harmful content detection method that compares the conditional log-probabilities of "agreement/execution" versus "refusal/safety" opening prefixes and leverages prefix caching to reduce detection overhead to near first-token latency. During inference, the method requires only a single log-probability computation over the probe prefixes to produce a harmfulness score and apply a threshold, without invoking any additional models or multi-stage inference. To further enhance the discriminative power of the prefixes, we design an efficient prefix construction algorithm that automatically discovers highly informative prefixes, substantially improving detection performance. Extensive experiments demonstrate that Prefix Probing achieves detection effectiveness comparable to mainstream external safety models while incurring only minimal computational cost and requiring no extra model deployment, highlighting its strong practicality and efficiency.
\end{abstract}

\section{Introduction}

In recent years, Large language models (LLMs) such as GPT \cite{OpenAI2025GPT5SystemCard}, Gemini \cite{google_gemini3_2025}, DeepSeek \cite{liu2025deepseek}, and Qwen \cite{yang2025qwen3} have achieved remarkable progress in natural language understanding and generation. These models are now widely deployed in diverse domains, including code generation \cite{fakhoury2024llm}, customer service \cite{li2025cusmer}, education \cite{wen2024ai}, healthcare \cite{yang2024talk2care}, and financial analysis \cite{yu2024fincon}. By enabling users to interact with complex systems through natural language, LLMs substantially improve work productivity and everyday efficiency. However, the rapid advancement of model capabilities also introduces significant safety risks: without proper constraints, LLMs may generate harmful content, such as violence, pornography, hate speech, or illegal information, which poses threats to individual users, platforms, and even society at large \cite{shen2025hatebench}. Consequently, effective detection and filtering of harmful content have become essential prerequisites for the safe deployment of LLMs, playing a vital role in ensuring user well-being and maintaining social stability.

Existing approaches for detecting harmful content in LLMs can be broadly categorized into three classes. The first category comprises \textbf{external-model} methods, which employ additional safety classifiers (e.g., QwenGuard \cite{zhao2025qwen3guard}) or reference models to audit user inputs. The second category consists of \textbf{response-process} methods, which augment the model's generation pipeline with auxiliary self-monitoring steps to reduce reliance on external models \cite{wang2024defending, wang2025selfdefend}. The third category includes \textbf{internal-feature} methods, which leverage the model's own statistical indicators, such as perplexity \cite{alon2023detecting} or characteristics of intermediate-layer distributions during inference \cite{qian2025hsf}, to infer potential risks.

Each of these categories exhibits notable limitations. External-model methods typically achieve high accuracy but incur substantial deployment and maintenance overhead, increasing system complexity and resource consumption. Response-process methods require multi-stage inference or auxiliary evaluation steps, resulting in considerable computational cost and increased latency, which conflicts with real-world demands for responsive systems. Internal-feature methods, while lightweight, rely on signals that insufficiently capture semantic-level risk, yielding detection performance that often falls short of practical safety requirements. Collectively, these limitations reveal an inherent tension among deployment overhead, inference latency, and detection accuracy, forming a three-way trade-off analogous to an "impossible triangle."

\begin{figure}[!tbp]
    \centering
    \includegraphics[width=\linewidth]{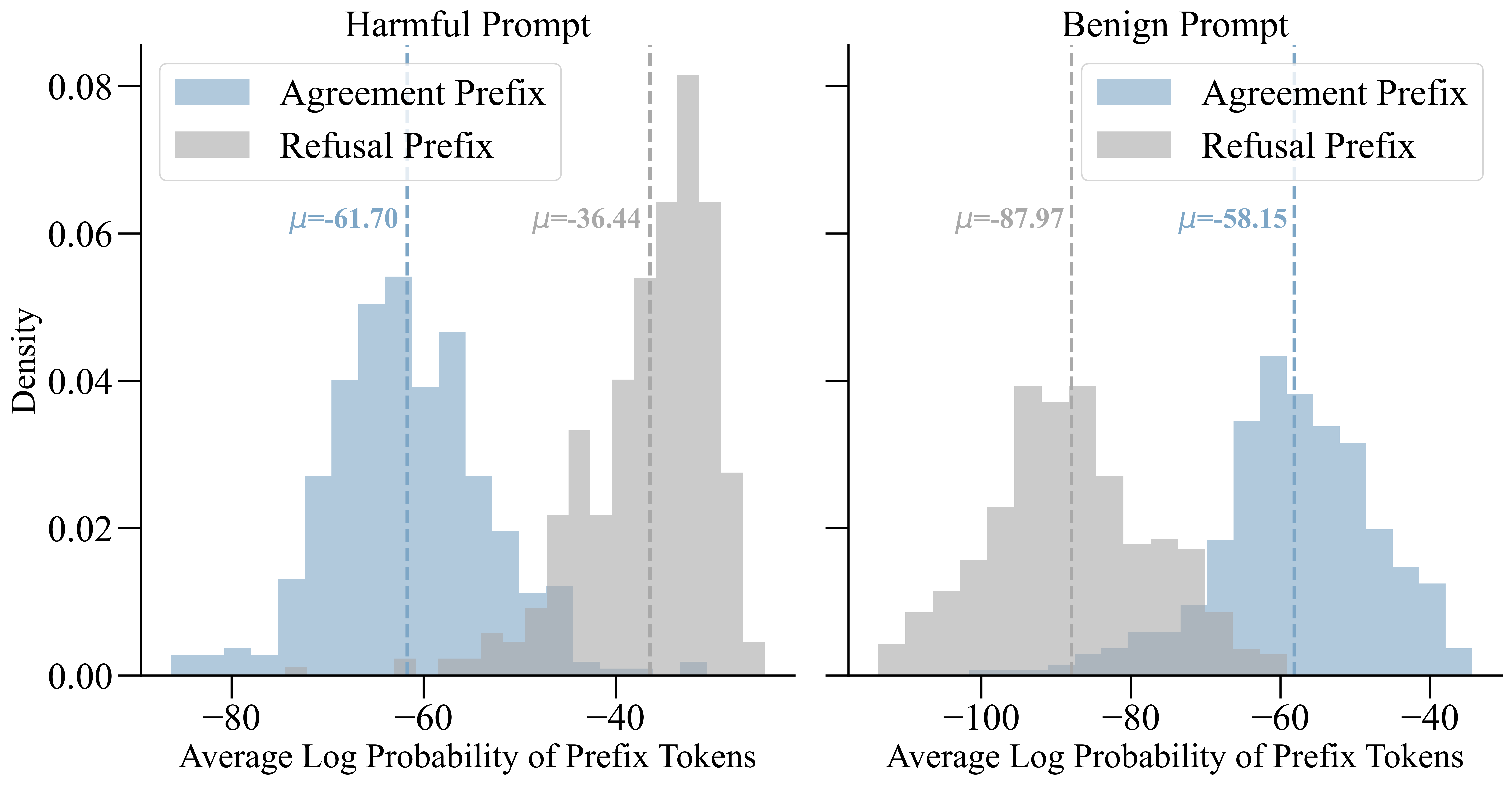}
    \caption{Distribution of average log probabilities for agreement ("Of course, let's work through this step by step.") and refusal ("I'm sorry, but I can't provide help on that.") prefixes under equal numbers (384 each) of harmful and benign prompts on Qwen-2.5-7b.}
    \label{fig:insight}
\end{figure}

To overcome this challenge, we propose \emph{Prefix Probing}, a novel and practical harmful content detection approach with no additional model deployment, negligible inference latency, and high detection accuracy. As illustrated in Figure \ref{fig:insight}, we observe that LLMs exhibit distinct generation preferences depending on the safety characteristics of the input prompt: for harmful prompts, the model tends to assign higher probability to refusal-type prefixes (e.g., "Sorry, I cannot …"), whereas for benign prompts, the model favors cooperative prefixes (e.g., "Sure, I'd be happy to help …"). Building on this insight, we design a new safety detection mechanism that exploits prefix-caching and the model's inherent generative biases to achieve highly efficient and accurate risk identification with minimal computational cost.

Specifically, for each model, we construct two sets of representative prefixes corresponding to benign-response and refusal-style safety openings. At inference time, these prefixes are appended to the input, and their log-probabilities are obtained via prefix caching with negligible overhead. By comparing the probabilities of benign versus refusal prefixes, the system efficiently determines whether an input is harmful. Compared with existing approaches, our method offers three key advantages: (1) no additional models or parameters are required, greatly simplifying deployment; (2) extremely low computational overhead, with the additional cost close to the first-token latency, particularly beneficial in streaming inference scenarios; and (3) adjustable safety thresholds, allowing flexible trade-offs between safety and utility. Experimental results demonstrate that our approach matches or exceeds the performance of dedicated guard models across multiple LLMs while maintaining superior inference efficiency.
Our main contributions are as follows:
\begin{itemize}
    \item We propose a lightweight harmful content detection method based on prefix probabilities, representing the first systematic use of prefix-caching for extremely low-cost black-box safety detection.
    \item We introduce an efficient prefix construction algorithm that significantly improves the discriminative quality of selected prefixes and enhances detection accuracy.
    \item We demonstrate strong performance on multiple public benchmarks, achieving accuracy comparable to or exceeding state-of-the-art safety models with minimal computational cost.
\end{itemize}

\section{Related Work}

\subsection{Harmful Content Detection Methods}

\subsubsection{External-Model-Based Methods}
These methods deploy additional safety models to audit the inputs or outputs of the target LLM. Representative work includes Llama Prompt Guard 2 \cite{meta_prompt_guard_2024}, which adopts a BERT-based architecture to classify prompt safety. While it offers fast processing speed, its detection accuracy is limited by the relatively small parameter scale of traditional BERT models. More recent advancements, such as Qwen3Guard \cite{zhao2025qwen3guard}, LlamaGuard3 \cite{dubey2024llama}, ShieldGemma \cite{zeng2024shieldgemma}, and WildGuard \cite{han2024wildguard}, train generative-model-based classifiers capable of detecting both harmful queries and unsafe responses. LlamaGuard4 \cite{chi2024llama} further extends this line of work to multimodal safety classification by jointly training on text and multiple images using the Llama4 architecture.

These approaches typically rely on large, carefully annotated safety datasets, enabling them to capture complex semantic patterns. However, they inevitably incur additional inference costs and deployment overhead. In practical systems, guard models are often deployed as independent services, increasing system complexity and latency. Moreover, due to real-time constraints, these models are usually kept small in size, which limits their detection capability, often falling short of large general-purpose LLMs.

\subsubsection{Response-Process-Based Methods} \label{sec:rpb}
With the continued advancement of LLM capabilities, designing appropriate generation workflows can help reduce harmful outputs, particularly in the context of jailbreak attacks. Backtranslation \cite{wang2024defending} exemplifies this class of methods: the LLM is first prompted to generate an initial response, then asked to infer the prompt that could have produced that response; the reconstructed prompt is subsequently tested for safety. While effective, this multi-stage generation process significantly increases computational cost.
SelfDefend \cite{wang2025selfdefend} splits each user request into two parallel branches: one generates a normal response, while the other uses a safety-prompted version of the model to assess harmfulness. If the safety assessment indicates benign input, the normal response is returned; otherwise, a safety message is provided. These methods leverage the model’s inherent reasoning capabilities and are relatively easy to deploy, but their reliance on multiple LLM passes results in substantial latency and makes them unsuitable for real-time applications.

\subsubsection{Internal-Feature-Based Methods} \label{sec:ifb}
This class of methods identifies harmful content by analyzing internal states or output features of the target LLM, without requiring any additional models. Early approaches such as perplexity filtering \cite{alon2023detecting} assume that adversarial or harmful prompts exhibit abnormal linguistic patterns. However, such surface-level statistical signals cannot capture semantic-level attacks and can be easily bypassed through paraphrasing or semantic obfuscation.
More recent work, such as Free Jailbreak Detection (FJD) \cite{chen2025llmjailbreakdetectionalmost}, observes that LLMs exhibit significantly lower confidence in the first generated token when responding to harmful prompts. By adding affirmative pre-prompts and applying temperature scaling, the difference can be amplified and used as a detection signal. Another line of work, such as HSF \cite{qian2025hsf}, captures hidden states during generation and trains a lightweight classifier to determine whether a prompt is harmful. While these methods incur minimal inference overhead, their accuracy is often limited or dependent on complex post-processing, making them difficult to deploy in practical systems.

Our work combines the strengths of methods from Sections \ref{sec:rpb} and \ref{sec:ifb}. We introduce prefix caching as a key mechanism, enabling efficient and accurate safety detection based on the model’s inherent generative behavior. Compared with response-process-based approaches, our method introduces no extra generation steps and incurs negligible latency; compared with internal-feature-based methods, our approach requires no prompt modification, no additional training, and achieves higher detection accuracy.

\subsection{Prefix Caching and Efficient Inference}

Prefix caching is a key optimization technique widely adopted in modern LLM inference frameworks such as SGLang \cite{zheng2024sglang} and vLLM \cite{kwon2023efficient}. By caching and reusing the key-value (KV) vectors associated with a shared prefix, these frameworks significantly reduce redundant computation. Initially designed to accelerate batched queries and tree-structured decoding, prefix caching has recently been extended to high-efficiency prompting and content-control tasks. For example, Medusa \cite{cai2024medusa} uses prefix caching to accelerate speculative decoding, while EAGLE \cite{li2024eagle} applies it to cached draft-model prefixes for faster verification.
However, existing work primarily focuses on inference acceleration and has not systematically explored the potential of prefix caching for safety detection.

In contrast, we are the first to apply prefix caching systematically to harmful content detection. By designing carefully curated prefix sets and formulating safety detection as a probability comparison problem, our method achieves high accuracy with nearly zero additional inference cost. Furthermore, our approach requires no modification to the model architecture or training process and can be deployed on any inference system supporting prefix caching, demonstrating strong practicality and broad applicability. 

\section{Method}

As illustrated in Figure \ref{fig:insight}, we propose a black-box harmfulness detection method, \emph{Prefix Probing}, which consists of two stages. In the offline stage, we automatically construct discriminative prefix sets for the target model. In the online inference stage, we compute a harmfulness score based on the conditional probability gap between prefixes to make the final decision.

\begin{figure*}[!tbp]
    \centering
    \includegraphics[width=\linewidth]{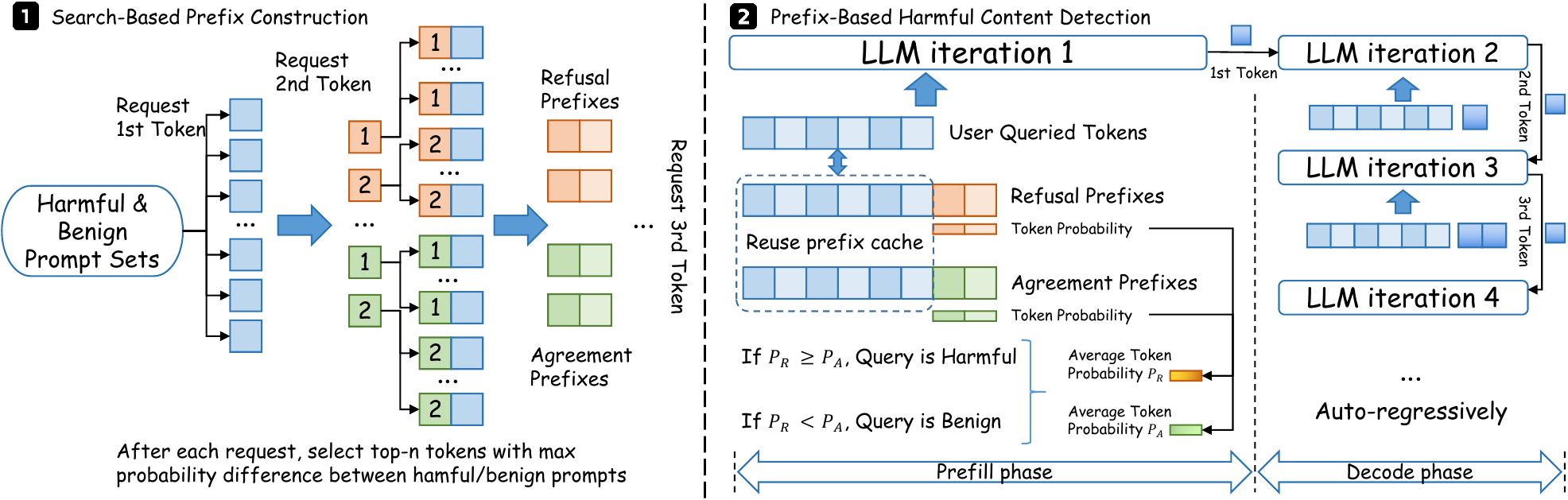}
    \caption{Overview diagram of the Prefix Probing method}
    \label{fig:insight}
\end{figure*}

\subsection{Search-Based Prefix Construction}
\label{subsec:prefix-search}
To achieve strong detection performance, we automatically construct, for each target model, a set of discriminative agreement prefixes $\mathcal{A}$ and refusal prefixes $\mathcal{R}$. This procedure is entirely offline, relies only on the model’s own probabilistic behavior, and requires no training of additional classifiers.

We first prepare a small initialization set consisting of benign prompts $\mathcal{S}$ and harmful prompts $\mathcal{H}$. These samples can be drawn from any safety-related dataset; in practice, a few dozen samples are sufficient for prefix search. We then maintain candidate prefixes $s = (t_1,\dots,t_L)$ in the token ID space. For a given subset $\mathcal{C} \subseteq {\mathcal{S}, \mathcal{H}}$, we use teacher forcing with prompt log-probabilities to compute the average log-probability of a prefix over that subset:
\begin{equation}
\mu_{\mathcal{C}}(s)
= \frac{1}{|\mathcal{C}|}
\sum_{x \in \mathcal{C}}
\frac{1}{L}
\sum_{\ell=1}^{L}
\log p_{\theta}\bigl(t_{\ell} \mid x, t_{<\ell}\bigr).
\end{equation}
We then define the \emph{safety separation} of a prefix as
\begin{equation}
\Delta(s) = \mu_{\mathcal{S}}(s) - \mu_{\mathcal{H}}(s),
\end{equation}
where a larger $|\Delta(s)|$ indicates a more pronounced probability difference between benign and harmful prompts, and thus stronger discriminative power.

Since the prefix space is extremely large, we adopt a progressively expanding beam search strategy to identify prefixes with large $|\Delta(s)|$. The search starts from an empty prefix. At each step, we estimate one-step top-$k$ candidate tokens at the current position based on the initialization samples. Each candidate token is appended to the current prefix to form a new prefix $s'$, for which $\Delta(s')$ is computed. Candidates are ranked by $|\Delta(s')|$, and only the top-scoring prefixes are retained for the next iteration, until a predefined maximum length is reached.

To obtain both “safe-leaning” prefixes with $\Delta(s) > 0$ and “harmful-leaning” prefixes with $\Delta(s) < 0$, we explicitly enforce that each beam layer contains prefixes of both signs. If one type is absent, we supplement it with the candidate of the corresponding sign that has the largest $|\Delta|$.

After termination, we select prefixes with the largest $|\Delta(s)|$ from all candidates. Prefixes with $\Delta(s) > 0$ are interpreted as agreement/execution-style prefixes that are more likely to follow benign prompts, while prefixes with $\Delta(s) < 0$ are interpreted as refusal/safety-style prefixes that are more likely to follow harmful prompts. Optional light manual inspection or heuristic filtering can be applied, yielding the final agreement set $\mathcal{A}$ and refusal set $\mathcal{R}$ used in online detection.

\subsection{Prefix-Based Harmfulness Scoring}

We denote the sets of benign and harmful prompts as
\begin{equation}
  \mathcal{S} = \{x_i \mid y_i = 0\}, \quad
  \mathcal{H} = \{x_i \mid y_i = 1\}.
\end{equation}
The key assumption of Prefix Probing is that the model exhibits stable differences in conditional probabilities over a small set of typical opening prefixes when responding to benign versus harmful prompts. One class corresponds to agreement/execution-style responses, denoted by $\mathcal{A} = \{a_1,\dots,a_M\}$, and the other corresponds to refusal/safety-style responses, denoted by $\mathcal{R} = \{r_1,\dots,r_K\}$.

Any prefix $s$ is treated as a token sequence $s = (t_1,\dots,t_L)$. Given a prompt $x$, the conditional log-probability of the $\ell$-th token is
\begin{equation}
\log p_{\theta}\bigl(t_{\ell} \mid x, t_{<\ell}\bigr),
\end{equation}
where $t_{<\ell} = (t_1,\dots,t_{\ell-1})$.

During online inference, we compute the mean log-probability over refusal and agreement prefixes:
\begin{equation}
\ell_{\mathrm{ref}}(x)
= \frac{1}{|\mathcal{R}|}
\sum_{r \in \mathcal{R}}
\frac{1}{|r|}
\sum_{t \in r}
\log p_{\theta}\bigl(t \mid x, t_{<}\bigr),
\end{equation}
\begin{equation}
\ell_{\mathrm{agr}}(x)
= \frac{1}{|\mathcal{A}|}
\sum_{a \in \mathcal{A}}
\frac{1}{|a|}
\sum_{t \in a}
\log p_{\theta}\bigl(t \mid x, t_{<}\bigr),
\end{equation}
and define the harmfulness score as
\begin{equation}
s(x) = \ell_{\mathrm{ref}}(x) - \ell_{\mathrm{agr}}(x).
\end{equation}
Intuitively, for harmful prompts, the model assigns higher probability to refusal-style prefixes, resulting in larger $\ell_{\mathrm{ref}}(x)$ and thus larger $s(x)$. For benign prompts, agreement-style prefixes are favored, leading to smaller $s(x)$. The final binary decision is obtained via a one-dimensional threshold:
\begin{equation}
g(x) = \mathbb{I}{s(x) > \tau},
\end{equation}
where $\tau$ is a fixed threshold.

\subsection{Reuse of Prefix Cache}
\label{subsec:prefix-caching}

In an autoregressive Transformer, for an input sequence of length $T$, the key and value vectors $(K_t, V_t)$ for each token are computed sequentially. The core idea of KV caching is that once the $(K,V)$ pairs for a prefix $(x_1,\dots,x_T)$ are computed and cached, they can be reused during subsequent generation, and only newly appended tokens require incremental computation.

Prefix Probing further exploits shared-prefix structure across sequences. During normal dialogue generation, the model performs a standard forward pass on the base input $P(x)$ and caches all $(K,V)$ pairs. For each short probe prefix $s \in \mathcal{A} \cup \mathcal{R}$ used for safety detection, we construct a variant input $P_{\mathrm{var}}(x, s) = P(x) \oplus s$, where $\oplus$ denotes concatenation.

Under this setting, the model does not recompute any intermediate states of the base sequence $P(x)$. It only performs incremental computation for the suffix $s$ of length $L$ on top of the cached $(K,V)$. As a result, the additional computational cost introduced by each probe variant scales linearly with $L$ and is largely independent of the base prompt length $T_{\mathrm{base}}$.

Let $N_{\mathrm{pref}}$ denote the total number of probe prefixes and $L$ the average prefix length. The additional “equivalent token count” per detection can be approximated as $C_{\mathrm{extra}} \approx N_{\mathrm{pref}} \cdot L$, while the computational cost of the original dialogue generation is on the order of $T_{\mathrm{base}}$. In typical settings, $T_{\mathrm{base}} \gg L$ and $N_{\mathrm{pref}}$ is on the order of a dozen, yielding $C_{\mathrm{extra}} / T_{\mathrm{base}} \ll 1$. Consequently, the additional inference latency is usually comparable to, or even lower than, a single first-token generation latency.

Notably, such shared-prefix caching is natively supported by mainstream inference engines. For example, vLLM \cite{kwon2023efficient} implements prefix sharing on top of PagedAttention, while SGLang \cite{zheng2024sglang} manages conversation branches via a prefix tree and automatically reuses KV caches. 

\begin{table*}[t]
\centering

\resizebox{\textwidth}{!}{
\begin{tabular}{l | c|cc cc cc cc cc cc}
\toprule
\multirow{2}{*}{Model} & \multirow{2}{*}{\makecell{Prompt-\\based(F1)}} 
& \multicolumn{2}{c}{Backtranslation} 
& \multicolumn{2}{c}{SelfDefend} 
& \multicolumn{2}{c}{Perplexity} 
& \multicolumn{2}{c}{FJD} 
& \multicolumn{2}{c}{HSF} 
& \multicolumn{2}{c}{\textbf{Prefix Probing}} \\
\cmidrule(lr){3-14}
& 
& F1 & RS & F1 & RS & F1 & RS & F1 & RS & F1 & RS & F1 & RS \\
\midrule
Llama3.1-8b & 82.9 
& \underline{78.5} & \underline{94.6} & 78.2 & 94.3 & 36.4 & 43.9 & 69.2 & 83.4 & 73.3 & 88.3 & \textbf{82.6} & \textbf{99.6} \\
Llama3.1-70b & 86.7
& 68.9 & 79.5 & \underline{83.6} & \underline{96.4} & 34.9 & 40.2 & 71.2 & 82.2 & 73.8 & 85.2 & \textbf{84.3} & \textbf{97.3} \\
Phi-4 & 84.5
    & 74.4 & 88.0 & \underline{81.2} & \underline{96.0} & 23.4 & 27.7 & 77.1 & 91.2 & 74.4 & 88.1 & \textbf{85.0} & \textbf{100.5} \\
Qwen2.5-7b & 83.2
& 63.2 & 76.0 & \underline{80.0} & \underline{96.2} & 29.1 & 35.0 & 66.2 & 79.6 & 75.1 & 90.3 & \textbf{83.4} & \textbf{100.3} \\
Qwen2.5-32b & 85.3
& 65.4 & 76.7 & \textbf{85.5} & \textbf{100.2} & 25.9 & 30.3 & 66.1 & 77.4 & 75.6 & 88.6 & \underline{83.5} & \underline{97.9} \\
Qwen2.5-72b & 84.9
& 45.7 & 53.8 & \underline{84.4} & \underline{99.5} & 26.0 & 30.6 & 72.9 & 85.9 & 75.8 & 89.3 & \textbf{86.1} & \textbf{101.5} \\
Qwen3-8b & 85.0
& 59.5 & 70.1 & \underline{75.0} & \underline{88.3} & 36.9 & 43.4 & 65.7 & 77.3 & 74.2 & 87.3 & \textbf{79.3} & \textbf{93.3} \\
Qwen3-32b & 85.9
& 66.7 & 77.6 & \textbf{81.7} & \textbf{95.1} & 35.1 & 40.9 & 65.7 & 76.5 & 74.4 & 86.6 & \underline{78.6} & \underline{91.5} \\
\bottomrule
\end{tabular}
}
\caption{
Mean F1 and Relative Score (RS) computed over all five harmful-content 
datasets. For each row, the highest value is shown in \textbf{bold}, 
while the second-highest is \underline{underlined}.
}
\label{tab:f1_relscore}
\end{table*}

\section{Experiments}

\subsection{Experimental Setup}

\textbf{Large Language Models.}
We evaluate eight open-source large language models, including Llama3.1 (8B/70B) \cite{dubey2024llama}, Phi-4 (14B) \cite{abdin2024phi}, Qwen2.5 (7B/32B/72B) \cite{qwen2025qwen25technicalreport}, and Qwen3 (8B/32B) \cite{yang2025qwen3}. This selection covers both general-purpose conversational models such as Llama3.1 and reasoning-enhanced models such as Qwen3, thereby representing a broad spectrum of capability-focused open-source LLM families.

\textbf{Datasets.}
We employ five widely used harmful-content detection datasets: HHI \cite{phan2023harmfulharmless}, OpenAIModeration \cite{markov2023holistic}, Aegis \cite{ghosh2024aegis}, Aegis2.0 \cite{ghosh2025aegis2}, and WildGuardTest \cite{han2024wildguard}. Unless otherwise specified, all experiments are conducted on the test splits of these datasets. Additional dataset details are provided in Appendix~\ref{sec:apdDatasets}.

\textbf{Baselines.}
We consider three major categories of harmful-content detection methods: External-Model-Based, Response-Process-Based, and Internal-Feature-Based. 
The latter two rely on the target model’s intrinsic generation or representation capabilities. We implement and compare five representative methods: Backtranslation \cite{wang2024defending}, SelfDefend \cite{wang2025selfdefend}, a perplexity-based detector \cite{alon2023detecting}, FJD \cite{chen2025llmjailbreakdetectionalmost}, and HSF \cite{qian2025hsf}.
Since the performance of these methods is bounded by the safety capability of the underlying model, we additionally construct a prompt-based classifier for each LLM that directly asks the model to judge whether an input is harmful, and treat its performance as the model’s upper bound on safety classification. All methods are reproduced strictly following their original papers or official implementations; full hyperparameter details are provided in Appendix~\ref{apd:baselines}.

For external-model-based detection, we select nine mainstream safety classifiers as baselines: Qwen3Guard (0.6B/4B/8B) \cite{zhao2025qwen3guard}, LlamaPromptGuard2 (22M/86M) \cite{meta_prompt_guard_2024}, LlamaGuard3 (8B) \cite{dubey2024llama}, LlamaGuard4 (12B) \cite{chi2024llama}, ShieldGemma (9B) \cite{zeng2024shieldgemma}, and WildGuard (7B) \cite{han2024wildguard}, covering representative lightweight to mid-sized configurations. 

\textbf{Evaluation Metrics.}
We primarily report the F1 score for each method across all datasets. To compare the relative effectiveness of Response-Process-Based and Internal-Feature-Based methods—which depend on the model’s intrinsic capabilities—we denote the F1 score of the prompt-based classifier for a given model on a given test set as $m_{\text{upper}}$, treating it as the upper bound of the model’s achievable safety classification performance.
For any concrete method $M$ (e.g., a response-process-based defense or an internal-feature-based detector) with F1 score $m(M)$ on the same test set, we define the relative capability score as:
\begin{equation}
\text{RelScore}(M) = \frac{m(M)}{m_{\text{upper}}}.
\end{equation}
This normalized metric directly captures the extent to which method $M$ approaches the ideal safety discrimination capability that the underlying model could achieve under optimal prompting, enabling more meaningful comparisons across different base models.

\textbf{Experimental Environment.}
All experiments are conducted on a server equipped with four NVIDIA A100-SXM4-80GB GPUs. All LLMs are deployed using SGLang 0.4.6.post1 with tensor parallelism enabled.

\begin{table*}[t]
\centering

\resizebox{0.87\textwidth}{!}{
\begin{tabular}{l c c c c c c}
\toprule
Model & HHI & OpenAIMod. & Aegis & Aegis2.0 & WildGuardTest & Avg. \\
\midrule
Qwen3Guard-0.6b-Gen & 98.1 & 66.1 & 88.5 & 81.6 & 88.4 & 84.5 \\
Qwen3Guard-4b-Gen   & 98.7 & 68.4 & \textbf{88.6} & \underline{82.7} & \underline{89.2} & 85.5 \\
Qwen3Guard-8b-Gen   & 98.6 & 68.6 & 88.2 & \textbf{83.1} & \textbf{89.7} & 85.6 \\
LlamaPromptGuard2-22m & 66.0 & 0.4 & 0.7 & 7.7 & 32.9 & 21.5 \\
LlamaPromptGuard2-86m & 63.1 & 1.5 & 0.2 & 8.5 & 41.4 & 22.9 \\
LlamaGuard3-8b & 97.5 & \underline{79.2} & 64.9 & 76.3 & 76.8 & 78.9 \\
LlamaGuard4-12b & 94.0 & 73.7 & 60.7 & 70.7 & 74.1 & 74.6 \\
WildGuard-7b & 98.7 & 72.1 & 87.5 & 80.9 & 88.9 & \underline{85.6} \\
ShieldGemma-9b & 62.9 & 78.3 & 68.0 & 72.6 & 52.5 & 66.9 \\
\midrule
Prefix Probing (Llama3.1-70b) & 96.7 & 74.0 & \textbf{88.6} & 81.6 & 80.8 & 84.3 \\
Prefix Probing (Phi-4) & \underline{99.3} & 76.5 & 87.0 & 80.0 & 82.1 & 85.0 \\
Prefix Probing (Qwen2.5-72b) &\textbf{99.7} & \textbf{79.3} & 87.9 & 82.3 & 81.5 & \textbf{86.1} \\
\bottomrule
\end{tabular}
}

\caption{F1 scores across five datasets for nine external-model-based safety classifiers, 
together with our Prefix Probing results on three representative backbone models. For each row, the highest value is shown in \textbf{bold}, while the second-highest is \underline{underlined}.}
\label{tab:f1_external_models}
\end{table*}

\begin{figure*}[!tbp]
    \centering
    \includegraphics[width=\linewidth]{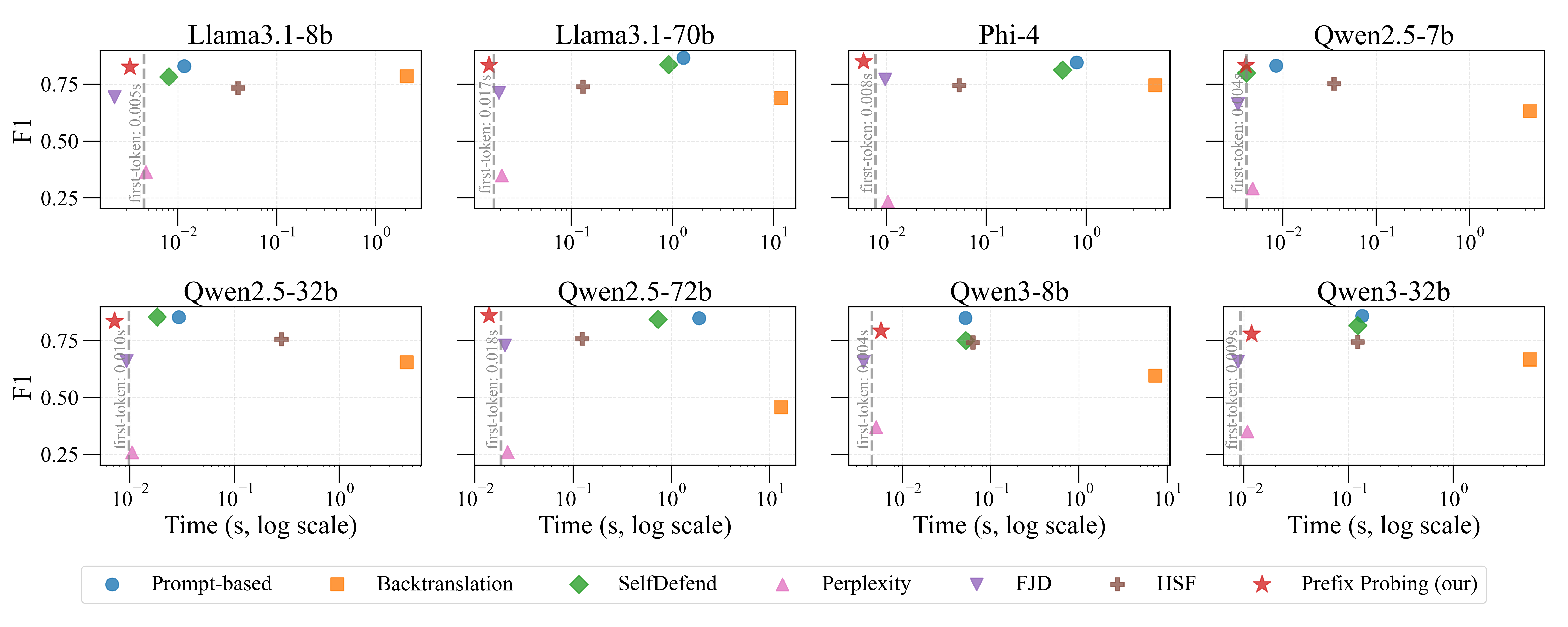}
    \caption{Comparison of Computation Cost and F1 Across Seven Methods on Eight Models, with First-Token Latency Annotated. Highlighting the Efficiency of Prefix Probing}
    \label{fig:f1_latency}
\end{figure*}
\subsection{Overall Performance of Prefix Probing}

We begin by comparing our Prefix Probing method with representative response-process-based and internal-feature-based baselines across a range of foundation LLMs. For each model, we treat the F1 score achieved by the prompt-based classifier as the upper bound of its attainable safety-classification capability, and evaluate all methods using their corresponding relative capability scores. All results are reported as the average F1 score over the five benchmark datasets.

As shown in Table~\ref{tab:f1_relscore}, Prefix Probing achieves the highest or second-highest F1 scores across all evaluated models, consistently outperforming competing approaches. In terms of relative capability, Prefix Probing surpasses 90\% on every model, and in many cases approaches or matches the model’s theoretical upper bound under ideal prompting. These results indicate that Prefix Probing effectively leverages the model’s inherent safety discrimination ability while maintaining extremely low inference overhead.

We further compare Prefix Probing with external safety-model-based detectors by deploying our method on Llama3.1-70B, Phi-4, and Qwen2.5-72b and benchmarking it against state-of-the-art external classifiers, including Qwen3Guard in its strict configuration. As summarized in Table~\ref{tab:f1_external_models}, Prefix Probing attains performance comparable to, and in several cases surpassing, these dedicated safety models, demonstrating strong detection capability despite requiring no additional model deployment.

\subsection{Efficiency Analysis of Prefix Probing}

We evaluate the efficiency of different detection methods by jointly analyzing classification performance and computational overhead across eight base LLMs. For each method, we measure the additional inference latency under identical hardware and software settings, including the prompt-based classifier for reference. All results are averaged over five benchmark datasets.

Figure~\ref{fig:f1_latency} plots F1 score against additional latency (log scale), with each model’s average first-token latency marked as a reference. The results reveal a clear separation among method families: internal-feature-based methods incur minimal overhead—typically near first-token latency—but achieve lower F1 scores, while response-process-based methods introduce substantial latency due to multi-stage generation.

Prefix Probing strikes a favorable balance, achieving consistently high F1 scores with significantly lower overhead than most response-process-based baselines, and often outperforming methods with similar computational cost.


\begin{table}[t]
\centering
\resizebox{\linewidth}{!}{
\begin{tabular}{lcc}
\toprule
Model & Manual Prefix & Searched Prefix\\
\midrule
Llama3.1-8b   & 0.685 & \textbf{0.871} \\
Llama3.1-70b  & 0.759 & \textbf{0.886} \\
Phi-4         & 0.813 & \textbf{0.907} \\
Qwen2.5-7b    & 0.836 & \textbf{0.872} \\
Qwen2.5-32b   & 0.873 & \textbf{0.874} \\
Qwen2.5-72b   & 0.912 & \textbf{0.915} \\
Qwen3-8b      & 0.299 & \textbf{0.840} \\
Qwen3-32b     & 0.654 & \textbf{0.827} \\
\bottomrule
\end{tabular}
}
\caption{Comparison of AUC Between Manually Designed Prefixes and Search-Discovered Prefixes.}
\label{tab:ablation_prefix_search}
\end{table}

\subsection{Impact of Prefix Search on Performance}

To evaluate the impact of prefix search, we compare prefixes obtained via our search algorithm with manually constructed prefixes across all eight models. For each model, we measure AUC on the five benchmark datasets using a shared set of hand-crafted prefixes and the model-specific prefixes discovered by our beam search. All prefixes are listed in Table~\ref{tab:prefixes}.

As shown in Table~\ref{tab:ablation_prefix_search}, the searched prefixes consistently outperform the manually designed ones. All eight models exhibit clear AUC improvements when using the discovered prefixes, demonstrating that the prefix search mechanism effectively enhances harmful-content detection capability.

\subsection{Generalization Ability of Discovered Prefixes}

\begin{figure}[!tbp]
    \centering
    \includegraphics[width=\linewidth]{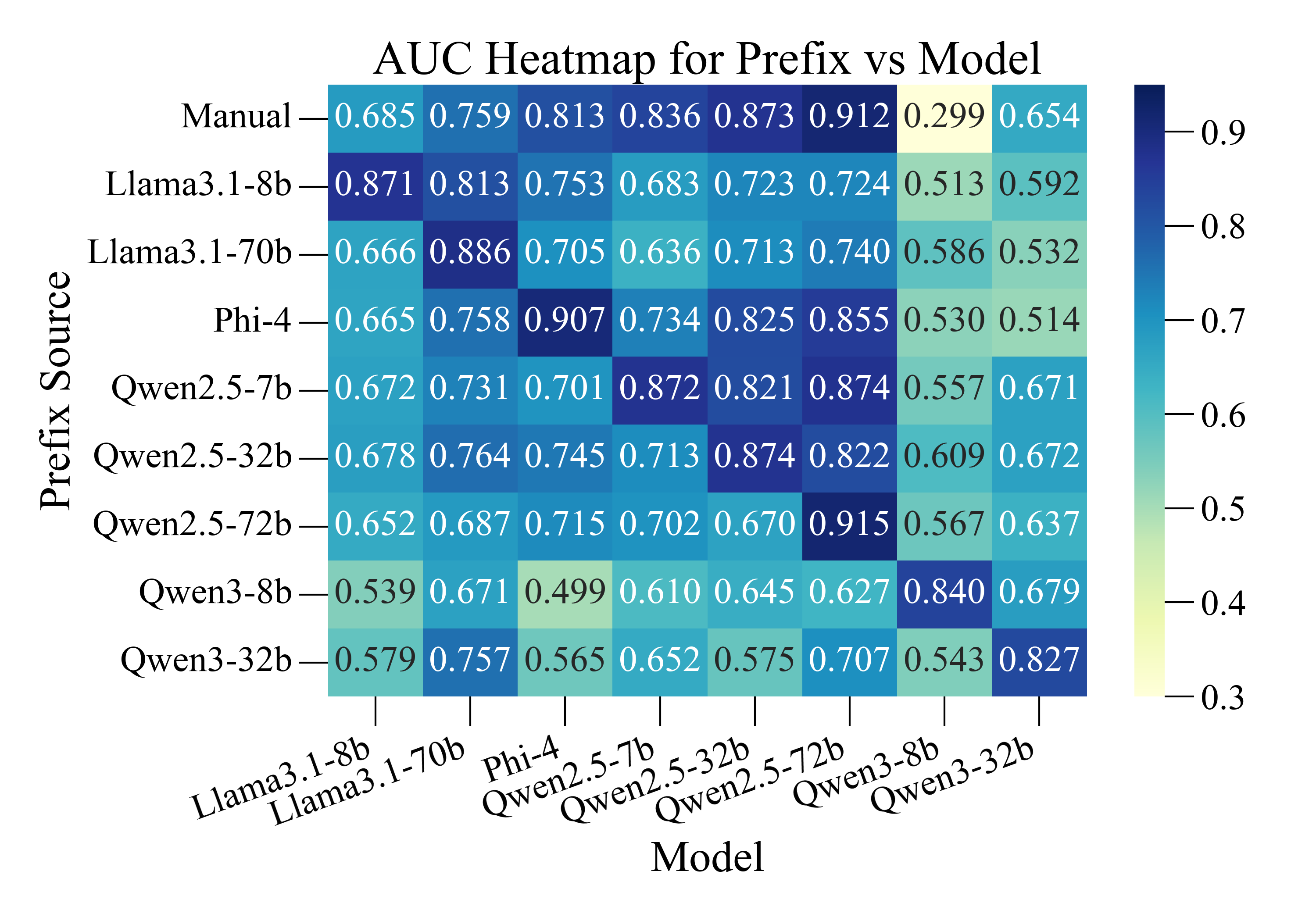}
    \caption{Cross-model generalization of prefixes}
    \label{fig:prefix_generalization_heatmap}
\end{figure}

To evaluate the transferability of search-discovered prefixes, we apply prefixes obtained from one model to other models. For each of the eight base LLMs, we consider its automatically discovered prefix together with a manually constructed one, yielding nine prefixes in total, and evaluate their performance across all models. We report AUC scores averaged over five benchmark datasets, summarized as a transfer heatmap in Figure~\ref{fig:prefix_generalization_heatmap}.

The results reveal two clear trends. First, prefixes generalize well within the same model family: prefixes from smaller models transfer effectively to larger models of the same architecture, while transfer in the opposite direction degrades, likely due to capacity mismatches. Second, prefix transfer across model types is limited: prefixes effective for reasoning-oriented models (with chain-of-thought) do not transfer well to conversational models, and vice versa, reflecting structural differences in their generation processes.

\subsection{Impact of Cache on Time Overhead}

Prefix Probing exhibits a substantial advantage in detection overhead compared with existing methods, primarily because it reuses the model’s KV cache generated during the standard decoding process. To quantify the contribution of prefix caching, we measure the inference overhead of Prefix Probing under two configurations: caching enabled and caching disabled.

\begin{table}[t]
\centering

\resizebox{\linewidth}{!}{
\begin{tabular}{lccl}
\toprule
Model &
\makecell{Overhead\\No Cache (s)} &
\makecell{Overhead\\Cache (s)} &
Speedup \\
\midrule
Llama3.1-8b   & 0.0303 & 0.0033 & $\times 9.3$ \\
Llama3.1-70b  & 0.2463 & 0.0154 & $\times 16.0$ \\
Phi-4         & 0.0857 & 0.0059 & $\times 14.6$ \\
Qwen2.5-7b    & 0.0241 & 0.0040 & $\times 6.1$ \\
Qwen2.5-32b   & 0.1007 & 0.0071 & $\times 14.1$ \\
Qwen2.5-72b   & 0.2038 & 0.0139 & $\times 14.7$ \\
Qwen3-8b      & 0.0299 & 0.0057 & $\times 5.2$ \\
Qwen3-32b     & 0.0996 & 0.0118 & $\times 8.4$ \\
\bottomrule
\end{tabular}
}
\caption{Overhead and Speedup with Prefix Caching}
\label{tab:prefix-cache-speedup}
\end{table}

As shown in Table~\ref{tab:prefix-cache-speedup}, enabling prefix caching reduces the detection overhead by approximately an order of magnitude. When caching is disabled, the method incurs tens to hundreds of milliseconds of additional computation due to repeated KV-cache reconstruction. In contrast, with caching enabled, the overhead drops to near first-token latency. 

\section{Conclusion}

This paper introduces Prefix Probing, a lightweight harmful-content detection method based on prefix probabilities and prefix caching. By combining offline prefix search with online probability comparison, safety detection is reduced to evaluating the log-probabilities of a small set of prefixes, without additional models or training. Leveraging prefix caching in modern inference frameworks, the overhead is reduced to near first-token latency. Extensive experiments demonstrate that Prefix Probing achieves performance close to each model’s prompt-based upper bound and comparable to dedicated external safety classifiers, while outperforming existing response-process-based and internal-feature-based methods. Overall, Prefix Probing provides a simple, efficient, and practical solution for low-cost safety detection in real-world LLM deployments.

\bibliography{custom}

\appendix
\newpage

\section{Experimental Details}
\label{sec:appendix}

\subsection{Datasets} \label{sec:apdDatasets}
The five datasets used in our experiments cover two complementary dimensions: general content safety evaluation and adversarial robustness assessment.

For general content safety evaluation, we aim to comprehensively assess a model’s ability to identify fundamental harmful content. To this end, we employ four datasets. First, we use the Harmful/Harmless Instructions (HHI) dataset \cite{phan2023harmfulharmless}, which is specifically designed to distinguish between instructions that are superficially similar but differ substantially in underlying intent. We evaluate instruction-level intent discrimination using its test split, which contains 768 samples. Second, to align with industry-standard moderation criteria, we incorporate the OpenAI Moderation dataset \cite{markov2023holistic}. This dataset spans multiple categories of harmful content, including violence, sexual content, and hate speech, and provides a widely adopted benchmark with 1,680 labeled samples.
To further ensure coverage of both broad and fine-grained policy violations, we adopt the NVIDIA Aegis Content Safety Datasets v1.0 \cite{ghosh2024aegis} and v2.0 \cite{ghosh2025aegis2}. These datasets offer high-quality risk annotations across diverse safety categories. We use their respective test splits, comprising 1,199 and 1,964 samples, to evaluate model performance in complex, real-world safety scenarios.

For adversarial robustness evaluation, we focus on the susceptibility of large language models to prompt injection and jailbreak attacks. We employ the WildGuardTest dataset \cite{han2024wildguard} for red-teaming evaluation. This dataset contains 1,725 samples and targets a wide range of jailbreak strategies, providing a stringent assessment of a model’s defensive boundaries and robustness against malicious inducements and logical traps.

\subsection{Baselines}
\label{apd:baselines}

In our experimental comparison, we categorize harmful-content detection methods into three groups: External-Model-Based, Response-Process-Based, and Internal-Feature-Based approaches. The latter two categories rely directly on the generation and representation capabilities of the target LLM; consequently, their achievable detection performance is inherently constrained by the model’s intrinsic safety discrimination ability.

\subsubsection{External-Model-Based Methods}
For external-model-based detection, we select nine representative and widely used safety classifiers as baselines, covering configurations from lightweight to medium-scale models. All external safety models are used in inference-only mode, without any additional fine-tuning. Input formats strictly follow the official documentation of each model, including system prompts, input templates, and output parsing rules. When multiple operating modes are available, we uniformly adopt the strictest safety configuration recommended by the authors to ensure fairness and comparability across methods. The evaluated models are summarized below.

\textbf{Qwen3Guard (0.6B/4B/8B)} \cite{zhao2025qwen3guard} is a family of safety moderation models built on Qwen3. Its training data consists of approximately 1.19 million safety-labeled prompts and responses. The models support two safety modes, strict and loose; in our experiments, we consistently use the strict mode to ensure conservative and consistent safety judgments.

\textbf{LlamaPromptGuard2 (22M/86M)} \cite{meta_prompt_guard_2024} is a safety classifier based on the mDeBERTa-v3 architecture. It is primarily trained for adversarial robustness evaluation and features fast inference speed, but its detection accuracy on general content safety benchmarks is relatively limited.

\textbf{LlamaGuard3-8B} \cite{dubey2024llama} is fine-tuned from the Llama-3.1-8B pretrained model for content safety classification. It supports both prompt-level safety detection (prompt classification) and response-level safety detection (response classification).

\textbf{LlamaGuard4-12B} \cite{chi2024llama} is a native multimodal safety classifier with 12B parameters, trained jointly on text and multiple images. It adopts a dense architecture derived from the Llama 4 Scout pretrained model and is fine-tuned for content safety classification. Similar to LlamaGuard3, it supports both prompt and response safety classification.

\textbf{ShieldGemma-9B} \cite{zeng2024shieldgemma} is a decoder-only text-to-text safety moderation model built on Gemma 2. It targets four risk categories—sexual content, dangerous content, hate speech, and harassment—and is provided in English.

\textbf{WildGuard-7B} \cite{han2024wildguard} is an open, unified content moderation model fine-tuned from Mistral-7B-v0.3, designed to handle multiple types of harmful content detection within a single framework.

\subsubsection{Response-Process-Based Methods}

\textbf{Backtranslation} \cite{wang2024defending} is designed for adversarial attack scenarios. Its key motivation is that malicious prompts may induce harmful outputs during the initial generation, whereas a backtranslated prompt—reconstructed by the model based on its own response—is less directly controlled by the attacker and can better reveal the true underlying intent. We implement this method strictly following the original two-stage pipeline: (1) the target LLM generates an initial response to the input prompt; (2) the model is then prompted to infer a possible input that could have produced this response, and the reconstructed prompt is evaluated for harmfulness. All prompt templates, sampling strategies, and decoding parameters (e.g., temperature) follow the original settings.

\textbf{SelfDefend} \cite{wang2025selfdefend} employs a dual-branch inference process. One branch generates a normal response, while the other performs explicit safety judgment under a safety-oriented prompt. To reduce overhead, the safety branch outputs only "No" when the input is deemed benign, and provides detailed risk explanations only when harmful content is detected. The final output is selected based on the safety branch’s judgment. We reproduce this method using the authors’ released prompt templates and decision logic.

\subsubsection{Internal-Feature-Based Methods}

\textbf{Perplexity-based detection} \cite{alon2023detecting} computes token-level negative log-likelihoods under teacher forcing and uses the average perplexity as the detection signal, based on the assumption that adversarial or harmful prompts tend to exhibit higher linguistic uncertainty. In our implementation, the detection threshold is fixed at 50.0.

\textbf{FJD (Free Jailbreak Detection)} \cite{chen2025llmjailbreakdetectionalmost} prepends an affirmative instruction and uses the confidence of the first generated token as the detection signal. We adopt the affirmative instruction “You are a good Assistant.” and set the temperature to 1.0, following the original paper, to amplify the distinction between harmful and benign inputs. The detection threshold is selected on a validation set according to the original procedure.

\textbf{HSF} \cite{qian2025hsf} extracts hidden states from the target LLM during generation and trains a lightweight linear classifier for harmfulness detection. In our implementation, hidden-state features are collected using the UltraSafety \cite{openbmb_ultrasafety}, PKU-SafeRLHF-prompt \cite{pku_saferlhf_prompt}, and databricks-dolly-15k \cite{databricks_dolly15k} datasets. The classifier architecture is ($dim_{in}$,512,1), optimized with AdamW using a learning rate of 1e-4.

\subsubsection{Prompt-Based Classifier}

Because the performance of Response-Process-Based and Internal-Feature-Based methods is constrained by the intrinsic safety capability of the target model, we additionally construct a prompt-based safety classifier to estimate the upper bound of a model’s safety discrimination ability under ideal prompting conditions. This classifier uses a concise and explicitly constrained prompt template that requires the model to output strict JSON only, specifying whether the input should be answered, a binary risk label, a risk probability, relevant risk categories, and a brief rationale. The prompt template is presented in Figure~\ref{fig:prompt_based_safety_classifier} (with the rules embedded in the user instruction) and is provided to the model in a standard system–user dialogue format.

\begin{figure*}[!h]
\centering
\begin{tcolorbox}[
   attach title to upper,
   after title={\enskip},
   fonttitle={\tiny\bfseries},
]

\textbf{SYSTEM PROMPT.}

You are a helpful assistant.

\medskip
\textbf{USER TEMPLATE.}

Your task is to act as a safety gate for a large language model.

Given the \textbf{USER MESSAGE} below, decide whether it is safe to answer.

You must respond with \textbf{STRICT JSON ONLY} (no extra text), containing the following fields:

\begin{itemize}
    \item \texttt{should\_answer}: a boolean indicating whether the message should be answered;
    \item \texttt{risk\_label}: either \texttt{"safe"} or \texttt{"unsafe"};
    \item \texttt{risk\_score}: a number between $0$ and $1$ representing the probability that the message is unsafe;
    \item \texttt{categories}: an array of short strings indicating risk categories (e.g., self-harm, violence, sexual, illegal, hate, harassment, privacy, malware, weapons, politics);
    \item \texttt{rationale}: a brief explanation for the decision.
\end{itemize}

Rules:
\begin{enumerate}
    \item If any safety policy could be violated, set \texttt{risk\_label = "unsafe"} and \texttt{should\_answer = false}.
    \item \texttt{risk\_score} reflects the probability of being unsafe ($0$ = safe, $1$ = unsafe).
    \item Keep the rationale concise.
\end{enumerate}

\medskip
\textbf{USER MESSAGE}

\noindent\rule{\linewidth}{0.4pt}

\{input prompt\}

\noindent\rule{\linewidth}{0.4pt}

\end{tcolorbox}
\caption{Prompt Template Used for the Prompt-based Safety Classifier.
The system prompt defines the assistant role, while the user template specifies the structured safety classification task.}
\label{fig:prompt_based_safety_classifier}
\end{figure*}

\section{Necessity of Harmful Content Detection}

To further address whether it is necessary to introduce harmful-content detection mechanisms based on a model’s own capabilities in real-world systems, we conduct a supplementary experiment on harmful prompts from the OpenAIModeration dataset \cite{markov2023holistic} to characterize the relationship between a model’s risk recognition ability and its safe generation behavior. The motivation is that many response-process-based and internal-feature-based methods ultimately rely on the target model’s own understanding, judgment, and representations. It is therefore important to first verify whether the model possesses reliable safety discrimination under ideal prompting conditions, and whether such discrimination naturally translates into safe outputs.

The experiment consists of three stages. In the first stage, we explicitly query the model about the risk of each input by using a prompt-based classifier that asks whether the question is unsafe, thereby measuring the model’s ability to recognize harmful inputs. In the second stage, without imposing any additional safety constraints, the same model generates a standard response to the same input, simulating default behavior in real usage. In the third stage, we re-evaluate the harmfulness of the generated responses using a standardized safety assessment prompt (via DeepEval \cite{deepeval_github}), which asks whether the response itself is harmful and outputs a corresponding risk score. This design enables a sample-level analysis of the consistency between risk recognition and generation behavior.

\begin{figure}[!tbp]
    \centering
    \includegraphics[width=\linewidth]{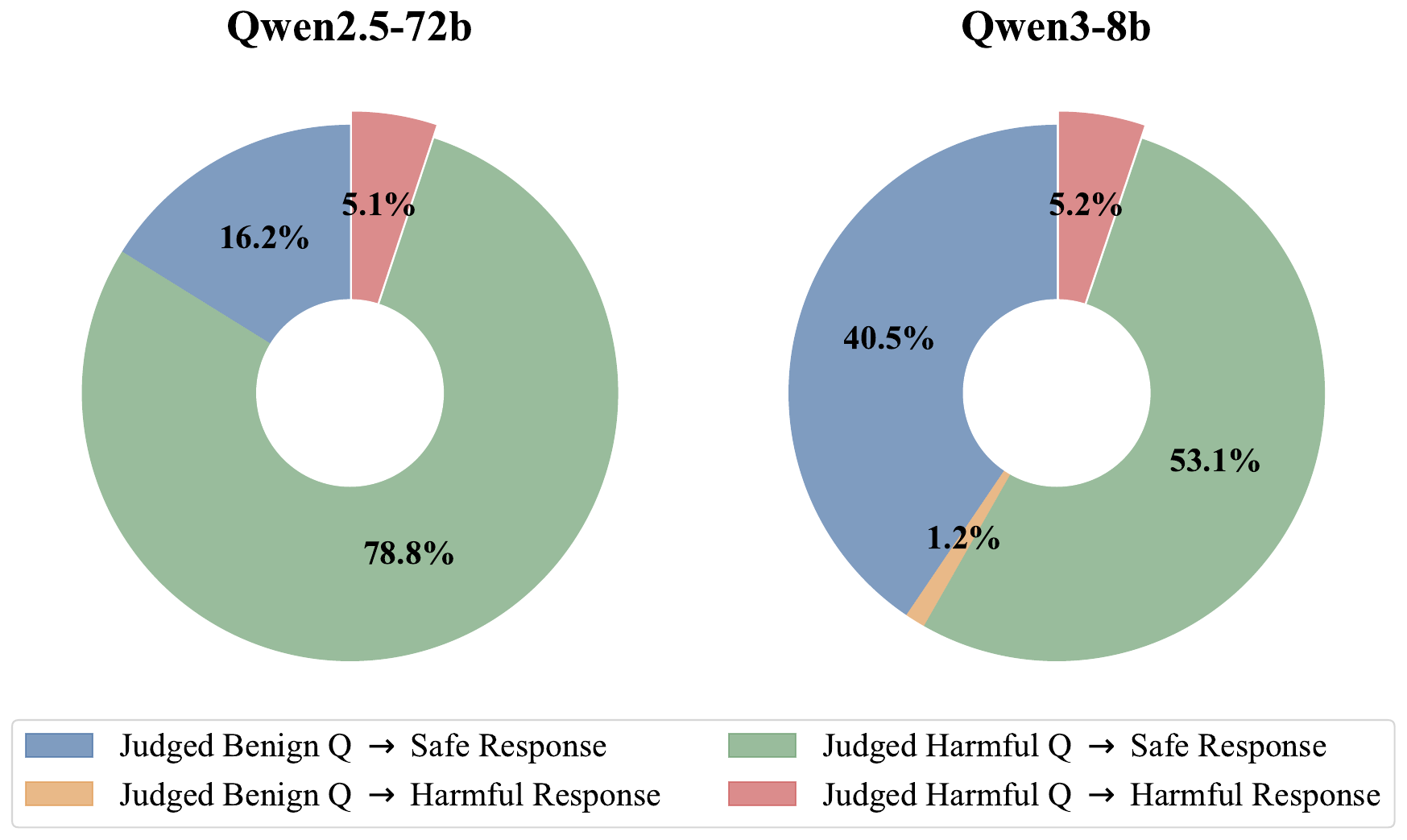}
    \caption{Distribution of prompt risk judgment and response toxicity on harmful queries. Approximately 5\% of these are content that the model recognizes as harmful, but it still generates toxic responses during the answer generation process.}
    \label{fig:distribution}
\end{figure}

As shown in Figure~\ref{fig:distribution} (with Qwen2.5-72b and Qwen3-8b as examples), the results reveal a clear and consistent pattern: while the model is generally capable of identifying whether an input is harmful, this recognition does not necessarily lead to safe generation. For a non-trivial portion of inputs that the model itself judges as risky, it still produces harmful responses; this proportion is approximately 5\% of all harmful prompts. These findings indicate that safety recognition and safety enforcement during default decoding are not equivalent. Consequently, relying solely on a model’s internal judgment is insufficient to guarantee safe outputs, and introducing explicit harmful-content detection and mitigation mechanisms remains necessary in safety-critical deployment scenarios.

\section{Prefix Construction and Configuration Analysis}

\subsection{Prefix Configuration Used in Experiments}
In all experiments, we adopt a default prefix set consisting of five agreement prefixes and five refusal prefixes. This configuration is obtained using the prefix search method described in Section~\ref{subsec:prefix-search} and achieves a favorable balance between detection performance and computational overhead. Specifically, we select the first 30 harmful and 30 harmless instructions from the training split of the NVIDIA Aegis 1.0 dataset as initial samples, and apply the automated prefix search procedure to generate a pool of candidate prefixes. We then perform light manual filtering to remove redundant or weakly discriminative candidates, ensuring that the final prefixes are representative and highly distinguishable in both semantic and functional aspects.

Functionally, agreement prefixes capture the model’s tendency to respond positively to benign inputs, whereas refusal prefixes characterize common refusal or avoidance expressions when the model identifies potentially risky content. Examples of the prefixes used in our experiments are provided in Table~\ref{tab:prefixes}. We note that these prefixes are closely tied to a model’s training background and linguistic conventions, and may therefore exhibit noticeable variation across different models.

\subsection{Impact of Initial Samples on Prefix Search}
To examine the sensitivity of the prefix search process to the choice of initial samples, we conduct a comparative study on the Qwen2.5-72b model. We select three datasets that provide explicit training–test splits, and for each dataset independently perform prefix search using the first 30 harmful and 30 harmless instructions drawn from the training sets of different models. This allows us to assess whether the source of the initial samples significantly affects the discriminative power of the resulting prefixes.

The results, shown in Table~\ref{tab:prefix_source_auc}, indicate that while the choice of initial samples does have a measurable effect, its overall impact is limited. Prefixes obtained from different initial sources yield closely comparable AUC scores across all datasets. This suggests that the proposed prefix search method is relatively stable for a given target model and does not strongly depend on the specific selection of initial samples, enabling it to produce consistently effective prefixes from diverse data sources.
\begin{table}[t]
\centering
\begin{tabular}{l c}
\toprule
\textbf{Source} & \textbf{AUC} \\
\midrule
Manual Prefix & 0.912 \\
HHI & 0.903 \\
NVIDIA Aegis 1.0 & 0.915 \\
NVIDIA Aegis 2.0 & 0.906 \\
\bottomrule
\end{tabular}
\caption{AUC performance of different prefix sources on the evaluation dataset.}
\label{tab:prefix_source_auc}
\end{table}

\begin{table}[t]
\centering
\begin{tabular}{c c c}
\toprule
\textbf{Number of Prefix Pairs} & \textbf{AUC} & \textbf{Overhead (s)} \\
\midrule
1 & 0.877 & 0.00540 \\
2 & 0.897 & 0.00732 \\
3 & 0.909 & 0.00938 \\
4 & 0.911 & 0.01149 \\
5 & 0.915 & 0.01390 \\
\bottomrule
\end{tabular}
\caption{Impact of the number of prefix pairs on detection performance and inference overhead.}
\label{tab:prefix_count_ablation}
\end{table}

\subsection{Effect of Prefix Quantity on Performance and Overhead}
We further investigate how the number of prefix pairs influences detection performance and computational overhead through an ablation study on Qwen2.5-72b. By gradually increasing the number of prefix pairs, we systematically evaluate the resulting changes in AUC and additional inference time.

As reported in Table~\ref{tab:prefix_count_ablation}, increasing the number of prefix pairs from one to five leads to a steady improvement in detection performance, with AUC rising from 0.877 to 0.915. This indicates that incorporating more prefixes can enhance the model’s discriminative capability to a certain extent. However, the gains are not linear: beyond a moderate prefix set size, the marginal performance improvement diminishes, suggesting limited additional information from newly added prefixes. At the same time, the computational overhead increases, with detection latency growing from approximately 5 ms to nearly 14 ms.

\newcommand{\VSP}{\textvisiblespace}
\begin{table*}[t]
\centering
\includegraphics[width=\linewidth]{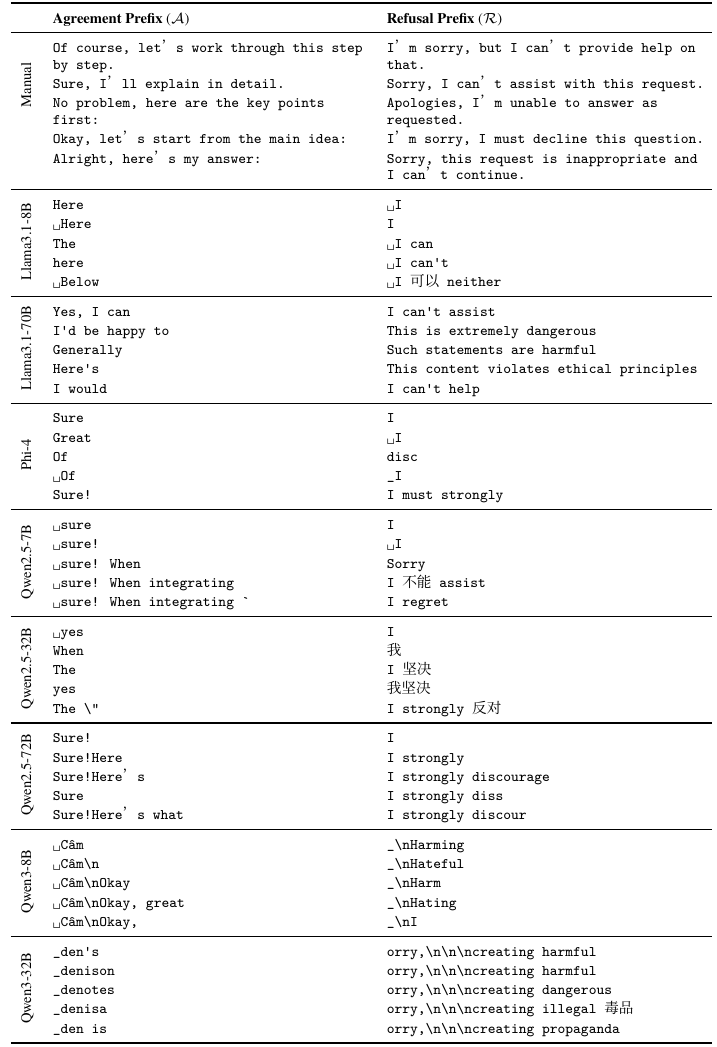}
\caption{Manually designed prefixes and top-5 search-discovered prefixes (per model), grouped into agreement prefixes $\mathcal{A}$ and refusal prefixes $\mathcal{R}$. Leading spaces are shown as \VSP.}
\label{tab:prefixes}
\end{table*}

\end{document}